\documentclass[lettersize,journal]{IEEEtran}

\usepackage{amsmath,amsfonts}
\usepackage{algorithmic}
\usepackage{algorithm}
\usepackage{array}
\usepackage[caption=false,font=normalsize,labelfont=sf,textfont=sf]{subfig}
\usepackage{textcomp}
\usepackage{stfloats}
\usepackage{url}
\usepackage{verbatim}
\usepackage{graphicx}
\usepackage{cite}
\usepackage{booktabs}
\usepackage[table]{xcolor} 
\usepackage{soul} 
\usepackage{makecell}
\usepackage{color}
\usepackage{tabularray}
\definecolor{darkgray}{rgb}{0.66, 0.66, 0.66}
\usepackage{hhline}
\usepackage{highlight}
\usepackage{etoolbox}
\usepackage{pgf}

\usepackage[colorlinks=true]{hyperref}
\usepackage[nameinlink]{cleveref}
\crefname{paragraph}{\S}{\S\S} 
\makeatletter
\DeclareRobustCommand{\textsupsub}[2]{{%
  \m@th\ensuremath{%
    ^{\mbox{\fontsize\sf@size\z@#1}}%
    _{\mbox{\fontsize\sf@size\z@#2}}%
  }%
}}
\makeatother

\begin{document}
\definecolor{highp}{HTML}{228b22}
\definecolor{lowp}{HTML}{fff7bc}
\newcommand*{\opacityp}{75}
\newcommand*{\minvalp}{0}
\newcommand*{\maxvalp}{1.0}
\newcommand{\gradientp}[1]{
    \ifdimcomp{#1pt}{>}{\maxvalp pt}{#1}{
    \ifdimcomp{#1pt}{<}{\minvalp pt}{#1}{
        \pgfmathparse{int(round(100*(#1/(\maxvalp-\minvalp))-(\minvalp*(100/(\maxvalp-\minvalp)))))}
        \xdef\tempa{\pgfmathresult}
        \cellcolor{highp!\tempa!lowp!\opacityp} #1
    }}
 }

\definecolor{highn}{HTML}{fff7bc}
\definecolor{lown}{HTML}{ef3b2c}
\newcommand*{\opacityn}{75}
\newcommand*{\minvaln}{-1}
\newcommand*{\maxvaln}{0}
\newcommand{\gradientn}[1]{
    \ifdimcomp{#1pt}{>}{\maxvaln pt}{#1}{
    \ifdimcomp{#1pt}{<}{\minvaln pt}{#1}{
        \pgfmathparse{int(round(100*(#1/(\maxvaln-\minvaln))-(\minvaln*(100/(\maxvaln-\minvaln)))))}
        \xdef\tempa{\pgfmathresult}
        \cellcolor{highn!\tempa!lown!\opacityn} #1
    }}
 }

\title{
Benchmarking the Robustness of Panoptic Segmentation for Automated Driving}

\author{Yiting Wang, Haonan Zhao, Daniel Gummadi, Mehrdad Dianati, Kurt Debattista and Valentina Donzella 
\thanks{Funded by the European Union (grant no. 101069576). Views and opinions expressed are however those of the author(s) only and do not necessarily reflect those of the European Union or European Climate, Infrastructure and Environment Executive Agency (CINEA). Neither the European Union nor the granting authority can be held responsible for them. UK and Swiss participants in this project are supported by Innovate UK (contract no. 10045139) and the Swiss State
Secretariat for Education, Research and Innovation (contract no. 22.00123) respectively.}
\thanks{The work was partially supported
by High Value Manufacturing CATAPULT. This research is
partially sponsored by the Centre for Doctoral Training to Advance the Deployment of Future Mobility Technologies (CDT FMT) at the University of Warwick}
\thanks{Yiting Wang, Haonan Zhao, Daniel Gummadi, Mehrdad Dianati, Kurt Debattista, and Valentina, Donzella are with WMG, University of Warwick, Coventry CV4 7AL, UK. Corresponding author: {\tt\small yiting.wang.1@warwick.ac.uk}}%
}


\maketitle




\begin{abstract}
Precise situational awareness is required for the safe decision-making of assisted and automated driving (AAD) functions. Panoptic segmentation is promising perception technique to identify and categorise objects, impending hazards, and driveable space at a pixel level. While segmentation quality is generally associated to the quality of the camera data, a comprehensive understanding and modelling of this relationship are paramount for AAD system designers.  Motivated by such a need, this work proposes a unifying pipeline to assess the robustness of panoptic segmentation models for AAD, correlating it with traditional image quality.
The first step of the proposed pipeline involves generating degraded camera data that reflects real-world noise factors. To this end, 19 noise factors have been identified and implemented with 3 severity levels. Of these factors, this work proposes novel models for unfavourable light and snow. After applying the degradation models, three state-of-the-art CNN- and vision transformers (ViT)-based panoptic segmentation networks are used to analyse their robustness. The variations of the segmentation performance are then correlated to 8 selected image quality metrics. This research reveals that: 1) certain specific noise factors produce the highest impact on panoptic segmentation, i.e. droplets on lens and Gaussian noise; 2) the ViT-based panoptic segmentation backbones show better robustness to the considered noise factors; 3) some image quality metrics (i.e. LPIPS and CW-SSIM) correlate strongly with panoptic segmentation performance and therefore they can be used as predictive metrics for network performance. The benchmark and code will be made available at \url{http://}
\end{abstract}

\begin{IEEEkeywords}
Automated driving, panoptic segmentation robustness, automotive image quality, noise factors.
\end{IEEEkeywords}

\section{Introduction}
Automotive camera data is commonly used in assisted and automated driving (AAD) systems to sense and interpret the vehicle's surroundings through the use of perception algorithms. The quality of the data and its relationship with perception algorithms' performance can have strong implications on the safety-critical decision-making processes used to navigate the environment. 

\begin{figure}[t!]
\centering
\includegraphics[width=.5\textwidth]{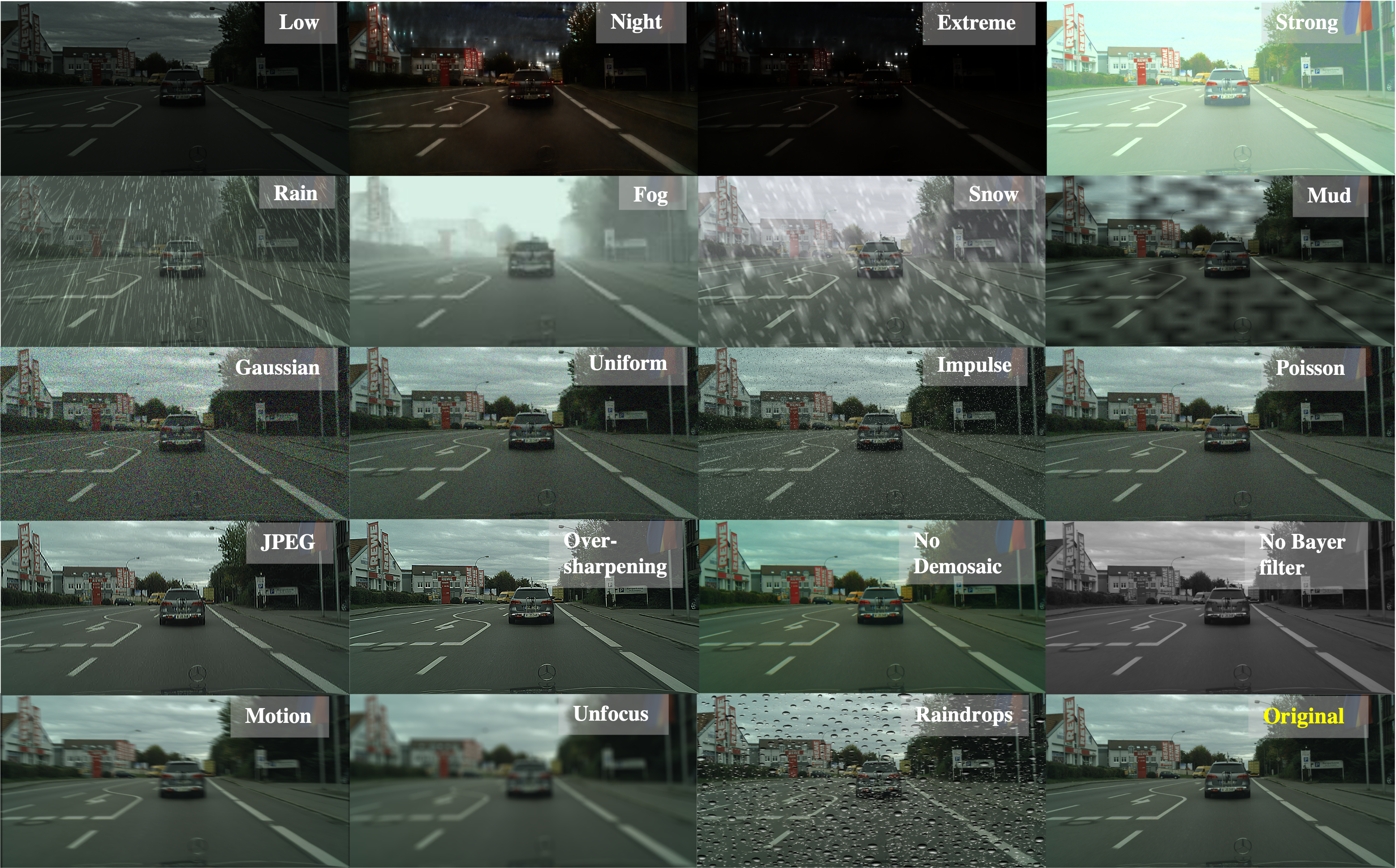}
\caption{Visual examples of the newly proposed Degraded-Cityscapes plus (D-Cityscapes+) with 19 types of degradation, from the top to the bottom, are categorised as unfavourable light, adverse weather, internal sensor noises, motion blur and distortion artefacts.}
\label{D-Cityscapes+}
\vspace{-5pt}
\end{figure}

Amongst various perception tasks, panoptic segmentation helps the AV identify countable objects (e.g. cars and pedestrians) in addition to background pixel categories (e.g. sky and grass) by segmenting pixels at the instance and semantic levels \cite{kirillov2019panoptic}. This granularity of information is essential for making well-informed and precise decisions and for responding to potential hazards in AAD functions. Neither object detection, semantic segmentation, nor instance segmentation can get such detailed scene information \cite{kirillov2019panoptic}. Panoptic segmentation produces accurate information on the unique shape and boundaries of each object via the classification of different instances of cars, pedestrians, bikes, etc. Despite the advantages of this technique, various data degradation factors in the real world might degrade its performance (see Fig. 1). Therefore, for the safety of AAD functions, it is essential to investigate panoptic segmentation robustness under degraded camera data.

Many researchers have investigated the robustness of perception tasks, such as object detection and semantic segmentation, using degraded sensor data. However, there is limited work that thoroughly considers camera degradation factors and implications on AAD panoptic segmentation \cite{hendrycks2019benchmarking,dong2023benchmarking,kamann2020benchmarking,wang2023semantic}. In addition, there is a lack of paired datasets covering a diversity of degradation factors~\cite{brophy2023review,8666747,ceccarelli2022rgb}. 
For instance, the previous robustness research disregarded dark scenarios, which are thought to be one of the primary causes of accidents
\cite{chebrolu2019deep,dong2020benchmarking,kamann2021benchmarking,ding2021perceptual,ceccarelli2022rgb,dong2023benchmarking}.
Due to the difficulty of capturing the degraded data in the real world, mainstream robustness research uses synthetic data, and these generated datasets might have qualitatively unconvincing results, Fig.~\ref{realism} \cite{chebrolu2019deep}. \textbf{There is a lack of systematic research considering the robustness of panoptic segmentation specifically for automated driving systems}.

To address the above-mentioned challenges, this work proposes a unified degradation impact pipeline, including 19 camera noise models, perception via panoptic segmentation, and result evaluation correlating image quality with perception quality, see Fig.~\ref{pipline}. Firstly, we propose an enhanced version of the previously degraded dataset  from~\cite{wang2023effect}. 
Then, a realistic synthetic degraded driving dataset called D-Cityscapes+ is introduced, which includes more noise factors, with varying models of fog and rain synthesis, and improved snow and light modelling. We also address the uneven distribution of degraded frames, i.e. our dataset contains the same number of degraded images under all degradation types. This work also proposes three types of generation models for unfavourable dark light conditions: low light, night light and extreme light (See Tab. \ref{light}). Three state-of-the-art panoptic segmentation models are selected in this work using six different architectures based on convolutional neural networks (CNNs) and Visual ViT-based backbones to analyse their robustness to degraded data. Furthermore, the robustness of perception is correlated to eight chosen image quality indexes, panoptic quality and the correlation between them. 

\begin{table*}[t]\centering
\begin{center}
\caption{Comparison of related methods in terms of perception tasks, datasets used, diversity of the degradation factors in terms of sensor noises(noises), image signal processing (ISP), compression (Jpeg), low-light, mud drop (Mud), the consideration of the severity levels (Severity) and the correlation study between the image synthetic image quality and the perception performance.}
\begin{tabular}{l l l c c c c c c c c c}
\toprule[1pt]
\textbf{Paper} & \small\textbf{Perception Task} & \small \textbf{Dataset} & \small\textbf{Noise} & \small\textbf{ISP}  & \small\textbf{Jpeg}  & \small\textbf{lowlight} & \textbf{Mud} & \small \textbf{Severity} & \small\textbf{Correlate}  & \small\textbf{AAD} \\
\hline
\cite{dong2020benchmarking} CVPR'2020   &  image classification & synthetic (paired) & \checkmark &  x & \checkmark & x & x & x  &x &x  \\ \hline
\cite{kamann2021benchmarking} IJCV'2021 &  semantic segmentation  & synthetic (paired) & \checkmark & x & \checkmark & x & x & x  & x & x \\ \hline
\cite{ding2021perceptual} T-ITS'2021 & image restoration & synthetic (paired) & \checkmark &   x & \checkmark & x & x & x & x & \checkmark \\ \hline
\cite{ceccarelli2022rgb} T-DSC'2022 &  object detection &  synthetic (paired)  & \checkmark & \checkmark & x & x & x & \checkmark & x & \checkmark \\ \hline
\cite{zendel2022unifying} CVPR'2022 &  panoptic segmentation & real-world (unpaired) & x &  x & x & \checkmark & x & \checkmark & x & \checkmark \\ \hline
\cite{wang2022performance} T-IV'2022 &  3D object detection & real-world (unpaired) & x &  x  &  x  & \checkmark & x & x & x & \checkmark \\ \hline
\cite{dong2023benchmarking} CVPR'2023 &  3D object detection & synthetic (paired) & \checkmark &  x  &  x  & x & x & \checkmark & x & \checkmark \\ \hline
\cite{xian2024towards} IJCV'2024 &  depth estimation & real-world (unpaired)  & \checkmark & x & \checkmark & x & x & \checkmark & x & x \\ \hline
\textbf{Ours} & \textbf{panoptic segmentation} & \textbf{synthetic (paired)} & \textbf{\checkmark} & \textbf{\checkmark} & \textbf{\checkmark} & \textbf{\checkmark} & \checkmark  & \textbf{\checkmark} & \textbf{\checkmark} & \textbf{\checkmark} \\ 
\bottomrule[1pt]
\end{tabular}
\end{center}
\label{robustness_review}
\end{table*}

This research is the first work to qualify and quantify the effect of 19 camera data degradation factors on panoptic segmentation for AAD systems. The main contributions of this work are: (\textbf{I}) a systematic benchmarking of the robustness of the CNN- and ViT-based panoptic segmentation architectures; (\textbf{II}) a new augmented dataset (D-Cityscapes+) with 19 types of degradation (47 types considering the severity levels) to boost future robustness research for AAD; (\textbf{III}) new noise models for unfavourable light and snow; (\textbf{IV}) a correlation of panoptic segmentation robustness with image quality metrics.

\section{Related Work}
This section introduces relevant work regarding the impact of automotive camera noise models, driving datasets embedding degradation factors, and panoptic segmentation.

\subsection{Impact of Camera Data Degradation}
Automotive camera noise factors are a widely investigated topic \cite{li2022analysis}. For example, 
Ceccarelli et al. discuss the common camera failures during the imaging process (i.e. lens, camera body, Bayer filter, image sensor, ISP) using the FMEA method and giving quantitative analysis via object detection \cite{ceccarelli2022rgb}. Dong et al. simulated common corruptions in cameras and Lidar for 3D object detection in autonomous driving \cite{dong2023benchmarking}. These researchers simulate the degraded paired datasets via simple picture editing that is not designed specifically for automotive cameras, leading to unsatisfactory fidelity for automotive applications, see Fig.~\ref{realism}. 
\begin{figure*}[t!]
\centering
\includegraphics[width=\textwidth]{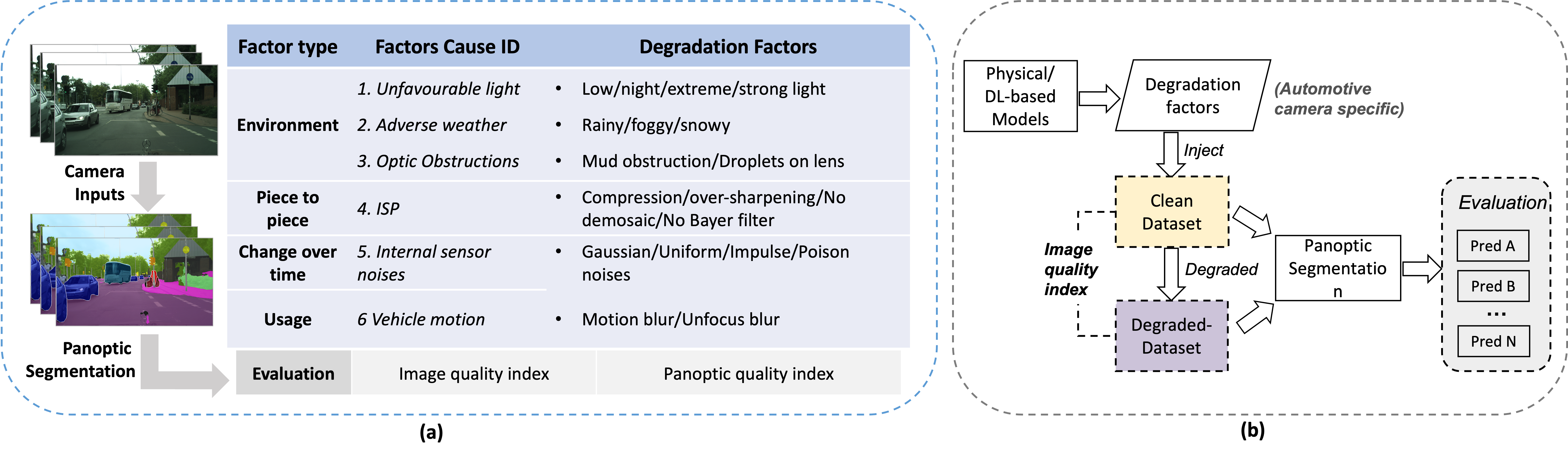}
\caption{The proposed unifying degradation impact pipeline consists of applying the noise factors to the clean dataset, panoptic segmentation, and evaluation. The 19 types of noise factors are included within the blue dotted box. }
\label{pipline}
\vspace{-5pt}
\end{figure*}
\begin{figure}[t!]
\centering
\includegraphics[width=.47\textwidth]{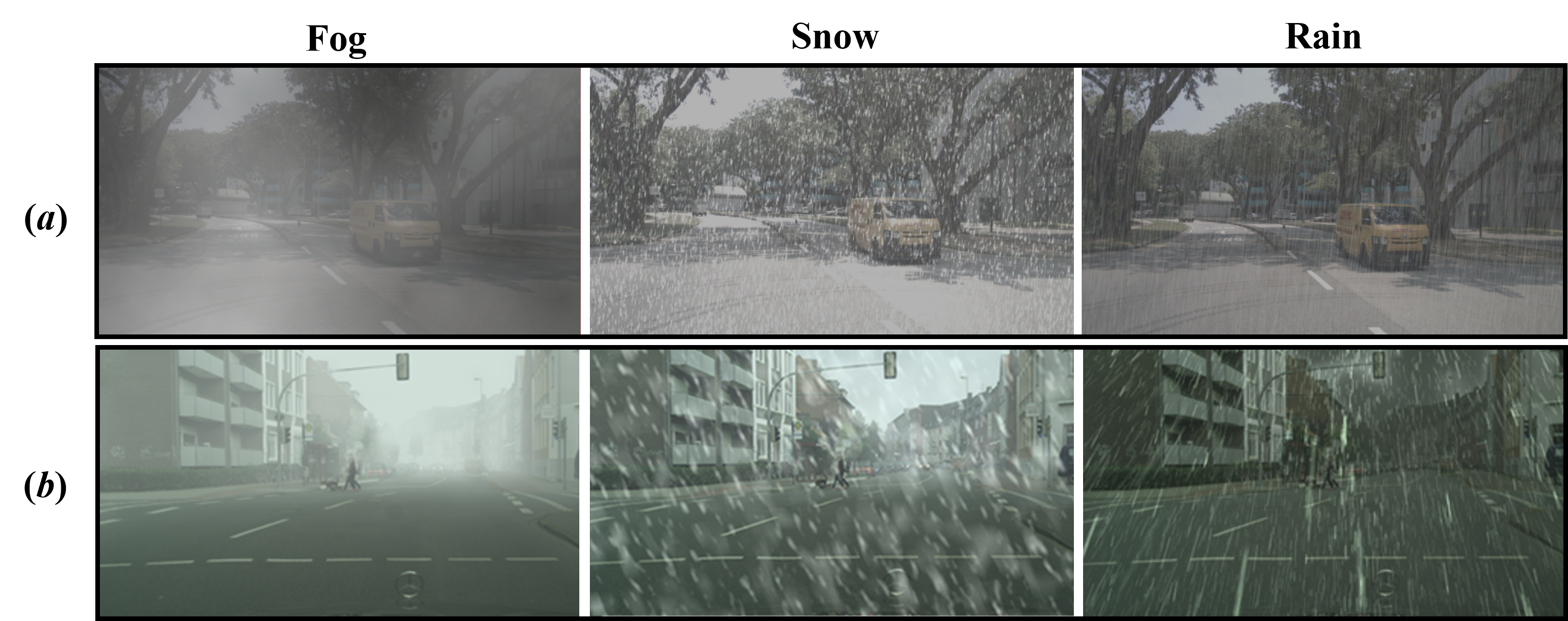}
\caption{Comparison of the adverse weather simulation used from (a) \cite{dong2023benchmarking} \textit{(CVPR'2023)} (directly using image editing tool, imgaug \cite{imgaug}) and (b) Ours (physical-based simulation methods referenced or modified from \cite{tremblay2021rain,sakaridis2018model,zhang2021deep}). \textit{We can see that: (1) The distribution of the fog and rain from (a) does not follow the physical rules with the depth information (2) The snow from (a) is evenly distributed with similar sizes, while ours mimics the random distribution and different sizes and directions of the snowflakes with a veiling effect.}}
\label{realism}
\vspace{-5pt}
\end{figure}
A summary of recent research on degraded camera data and the relationship with different perception tasks is in Tab.~\ref{robustness_review}. As indicated, there are a few works related to perception tasks specifically in automotive, such as image classification, object detection, and segmentation~\cite{dong2020benchmarking,dong2023benchmarking,wang2022performance,kamann2020benchmarking,wang2023semantic,xian2024towards}. Most of these works focus on natural looking images; for example, 15 types of corruption are synthesised in the ImageNet-C dataset~\cite{hendrycks2019benchmarking}. 
As the robustness of perception tasks is crucial for the safety of real-world applications, research is also emerging on the robustness of automotive camera data, especially under adverse weather \cite{bijelic2018benchmarking,8666747,brophy2023review}. 
For instance, Wang et al. generated 11 types of image degradation and tested the impact of one panoptic segmentation model \cite{wang2023effect}. 
Differently from related previous work, this paper: \textbf{1)} considers a wider range of noise factors, including unfavourable light conditions and pollutant particles (mud and stain) on the camera lens; textbf{2)} uses state-of-the-art noise models that are designed specifically for automotive applications to reduce the simulation to real-world (Sim2Real) gap for fairer robustness validation; \textbf{3)} analyses qualitatively and quantitatively the effect of noise model at a pixel level, including semantic and instance meaning. 

\subsection{Driving Datasets embedding Degradation Factors}
Various degradation-driven benchmarking datasets are available to facilitate the robustness of research and to improve the algorithm performance under degradation conditions; they can be divided into real-world captured datasets and synthetic augmented datasets. Many real-world datasets are captured with labels for specific tasks. For example, the WildDash1 and WilDDash2 datasets are used for benchmarking hazardous conditions for semantic segmentation and panoptic segmentation, respectively \cite{zendel2018wilddash,zendel2022unifying}. The ACDC dataset is captured under night, foggy, rainy and snow conditions \cite{sakaridis2021acdc}. BDD100K datasets contain additional conditions under dusk, overcast and cloudy weather \cite{yu2020bdd100k}. 

\textbf{Limitations of existing real-world driving datasets} \textbf{1)} Lack of labels and paired data for panoptic segmentation under various noise factors. \textbf{2)} Uneven distributions of noise factors, resulting in reduced generality and reliability, e.g. in BDD100k only 0.2\% of the images are collected in foggy conditions \cite{yu2020bdd100k}. 
\textbf{3)} Scarcity of extreme degradation levels \cite{zendel2022unifying,mușat2021multi}. 
As an alternative, image synthesis methods are proposed to use physical models \cite{hu2019depth,garg2007vision,tremblay2021rain,sakaridis2018semantic,zhang2021deep}, deep learning-based methods \cite{sakaridis2018model,sakaridis2019guided,mușat2021multi,zhang2021deep} and the virtual environments \cite{dosovitskiy2017carla,sun2022shift,gaidon2016virtual} to generate camera datasets including augmented degraded camera images. Except for some off-the-shelf generation models, challenges exist in generating datasets with adequate quality for adverse weather and unfavourable light conditions. Platforms such as CARLA might support the option of generating various environmental conditions using the embedded functions, but they have very limited image fidelity tailored specifically for automotive camera degradation compared with the real-world captured ones. Traditional techniques, on the other hand, involve complicated models and hand-crafted parameter formulations. Deep learning (DL)-based algorithms, which offer more flexibility, have limitations as they are dataset-dependent and lack simple solutions to regulate the severity of degradation. Additionally, the performance of DL-based approaches, particularly GAN-based methods \cite{zhu2017unpaired}, is inconsistent (e.g. failing to capture structural integrity) in some cases. Therefore, for the degraded data generation in this research, a combination of both traditional and DL-based approaches is selected, with the aim of generating a degraded dataset with better quality.

\subsection{Panoptic Segmentation}
State-of-the-art DL-based panoptic segmentation models can be divided into top-down, bottom-up, single path, and other techniques \cite{kirillov2019panoptic_FPN,xiong2019upsnet,mohan2021efficientps,cheng2020panoptic,yang2019deeperlab,li2021fully,li2022panoptic,jain2023oneformer,li2023mask}. Well established methods fuse instance and semantic segmentation information with the existing popular object detectors. For example, the feature pyramid networks are used in the panoptic FPN and the UPSNet to improve multiscale feature learning ability \cite{kirillov2019panoptic_FPN,xiong2019upsnet}. The lightweight Efficient Net is used in the EfficientPS to reduce the number of parameters  \cite{mohan2021efficientps}. The atrous convolution and dilated convolutions are used in the Panoptic Deeplab and DeeperLab to capture multiscale context information \cite{cheng2020panoptic,yang2019deeperlab}. In addition to the convolutional neural network (CNN)-based architectures, recently, ViT-based architectures have been proposed with improved performance, benefiting from self-attention mechanisms and long-range correlation learning ability~\cite{li2022panoptic,cheng2021per,li2023mask,jain2023oneformer,cheng2022masked,wang2021max}. For example, 
OneFormer proposes a universal image segmentation approach that can achieve better performance on multitasks (e.g. semantic, instance and panoptic segmentation) by training a single model with a single dataset compared with the existing methods, which require multi-usage of resources to train different single tasks \cite{jain2023oneformer}. 
In this research, we evaluate the impact of different degradation factors for panoptic segmentation using both the CNN-based and the ViT-based architectures with different backbones to compare their robustness. 

\section{Analysis of Selected Noise Factors}
In the context of this paper, \emph{degradation factor} is defined as any external (e.g. adverse environmental conditions) or internal factors (e.g. electronic noise) that may influence the quality of the generated sensor data, with a specific focus on the effects of such factors on the perception algorithms' performance for automated driving. Reliable evaluation of image quality and sensor perception robustness necessitates a comprehensive assessment of noise models. 
Using the P-diagram as a guiding tool, four categories of noise factors from the existing five are selected, i.e. piece-to-piece, change over time, usage, and environment~\cite{li2022analysis}. The fifth, system interactions, is considered out of the scope of this paper.
Within these 4 groups, a list of seven sub-categories is identified, as detailed in the following section, see, Fig.~\ref{pipline}, and a total of 19 noise factors are selected for this research (see Tab. \ref{simulation_table} for the implementation details). The selection of the 7 factors cause IDs are the currently most commonly seen noise factors impacting perception in AAD, and from which 19 specific noise factors that have the potential to be synthetically rendered are selected.

\noindent\textbf{Off-the-shell Models} Many common image noise models have been recently implemented based on relatively mature theoretical physical models; therefore, some well-established off-the-shell models have been used to generate some of the identified noise factors (cause factor ID = 1,4,5,6): strong light \cite{michaelis2019dragon}, Droplets on lens \cite{ceccarelli2022rgb}, JPEG Compression \cite{imgaug}, Over-sharpening \cite{imgaug}, No Bayer filter \cite{ceccarelli2022rgb}, internal sensor noises \cite{imgaug}, motion blur \cite{imgaug} and unfocus blur \cite{imgaug} (\textit{see supp. materials for details)}. Apart from these, noise factors that are designed specifically for automotive applications in mud obstructions on the camera lens and No Demosaic (Bayer data synthetic) (cause factor ID = 3,4) are also chosen in this application \cite{li2022analysis,chan2023raw}. As for the weather condition (cause factor ID = 2), since there are several augmented multi-weather cityscape datasets available, we choose the most suitable ones for the rainy and foggy conditions based on the reasons listed below \cite{mușat2021multi,sakaridis2018semantic,sakaridis2018model,zhang2021deep,hu2019depth,tremblay2021rain}. \textbf{\textit{1) Rain Model}}. Rain Cityscapes follows the physical rules considering the depth information with a rain layer and a fog layer \cite{hu2019depth}. However, the method fails to consider the photometry of the rain streak or the fact that the droplets on the lens will need a larger field of view from cameras. A new rain rendering method is proposed, which has the advantage of pre-defining the required rain rate (mm/hr) and producing more vivid rain streaks for generating realistic rainy images, therefore, chosen as the rain model \cite{tremblay2021rain}.   
\textbf{\textit{2) Fog Model}}. 
Foggy Cityscapes \cite{sakaridis2018model} uses the scattering model to augment the foggy images. As the original depth map provided from the dataset is incomplete and discontinuous (with random ``holes'' that missing the depth values), the depth denoising, completion, and guided filter are used to obtain the final transmission map \cite{sakaridis2018model}. However, the simple filter method fails to capture the boundaries between different semantic objects, which leads to invalid depth guidance in the simulated fog physical model. An improved version of Foggy Cityscapes-DBF is therefore proposed \cite{sakaridis2018semantic} with a dual-reference cross-bilateral filter for better adherence to semantic boundaries in the scene and hence chosen as the fog model in this research. 

\textit{The implementation details are shown in the Tab.} \ref{simulation_table}. The following sections explain the implementation of each of the 19 considered noise factors, including a more detailed description of the newly introduced noise models (cause factor ID = 1) related to unfavourable light and improved snow models. 

\begin{table*}[t]\centering
\begin{center}
\caption{Details in terms of the category ID, degradation factor names, synthetic implementation steps and configurations for the selected noise factors. }
\label{simulation_table}
\begin{tabular}{l l p{7cm}<{\raggedright} p{6cm}<{\raggedright}}
\toprule[1pt]
\textbf{ID} & \textbf{Degradation} & \small\textbf{Implementation Details } & \small \textbf{Configs. } \\
\hline

1 & Low light & {Apply the inverse of the DL-based curve estimation image enhancement method (EC-Zero-DCE \cite{zhou2022lednet})}  & The pre-trained model from EC-Zero-DCE \cite{zhou2022lednet} \\ \hline

1 & Night light & Retrain the CyCleGAN network \cite{zhu2017unpaired} with the BDD100K dataset \cite{yu2020bdd100k} & PyTorch with 2k epochs for training \\ \hline

1 & Extreme light & Apply the combination of the model of CycleGAN \cite{zhu2017unpaired} + EC-Zero-DCE \cite{zhou2022lednet} & Apply low-light model effect into the nightlight model \\ \hline

1 & Strong light & Apply the Python package from Imagecorruptions \cite{michaelis2019dragon} & Pre-defined brightness corruptions severity levels = \{1,3,5\} \\ \hline

2 & Rainy & Apply rain rendering method \cite{tremblay2021rain} with a physical simulator and accurate rain photometric modelling & Three pre-defined severity levels with rain rate (mm/hr) = {50,100,200} \\ \hline

2 & Foggy & Apply Foggy-DBF datasets with atmospheric scattering model from \cite{sakaridis2018model}  & The values of the attenuation coefficient = {0.005, 0.01, 0.02} with the visibility ranges(m) = \{600,300,150\} \\  \hline

2 & Snowy & Improve the Snow Cityscapes \cite{zhang2021deep} by 1) synthesising different snow masks on all 500 cityscapes validation images 2) adding the veiling effect \cite{chen2020jstasr}  & Snow mask originally generated using Photoshop from \cite{zhang2021deep} and divided into small, medium, large \\ \hline

3 & Mud obstruction & Apply the mud model from \cite{li2022analysis} with generated random mud masks and cv2 modules to add the mud and spain into the images  & Three levels of severity with kernel sizes \{12, 24, 36\} and intensity 0.7 \\  \hline

3 & Droplets on lens & Apply the PIL library to blend the raindrops into the image to simulate the droplets on lens on camera lens  & The raindrop masks from \cite{ceccarelli2022rgb} with severity level = \{2,3,4\} \\  \hline

4 & Compression & Apply imgaug.augmenters.arithmetic.JpegCompression(strength) in the imgaug library \cite{imgaug}  & Three different compression levels = \{20, 50, 80\} corresponding to three JPEG compression rates \{74.94\%, 58.80\%, 42.20\%\}\\ \hline

4 & Over-sharpening & Apply imgaug.augmenters.Sharpen(alpha, lightness) in the imgaug library \cite{imgaug} & Three levels of severity with alpha =\{0.25,0.5,0.75\} \\ \hline

4 & No Demosaic & Map RGB into RGGB pattern with Bayer colour-filled \cite{ceccarelli2022rgb} & Three channel same size (h, w)\\ \hline

4 & No Bayer filter & Apply RGB-to-Grayscale with numPy array & The scaling factor = \{0.2989, 0.5870, 0.1140\}  \\ \hline

5 & Gaussian noise & Apply imgaug.augmenters.imgcorruptlike.GaussianNoise() in the imgaug library \cite{imgaug} & Three pre-defined levels of severity =\{1,3,5\} \\ \hline

5 & Uniform noise & Apply mathematical model, CV2 and numPY to generate random distributed noises following \cite{imgaug} & Three Severity levels with mean value = 0 and standard deviation = \{25, 50, 75\}  \\ \hline

5 & Impulse noise & Apply imgaug.augmenters.arithmetic.ImpulseNoise(severity) in \cite{imgaug} & Three pre-defined levels of severity =\{1,3,5\}\\ \hline

5 & Poisson noises & imgaug.augmenters.arithmetic.AdditivePoissonNoise(lambda) in \cite{imgaug} & Three levels of severity with lambda =\{5,10,15\} \\ \hline

6 & Unfocus Blur & Apply imgaug.augmenters.imgcorruptlike.DefocusBlur() from \cite{imgaug} & Three pre-defined levels of severity =\{1,3,5\}\\  \hline

6 & Motion blur &  imgaug.augmenters.imgcorruptlike.apply\_motion\_blur() \cite{imgaug} & Three pre-defined levels of severity =\{1,3,5\} \\ \hline
\end{tabular}
\end{center}
\end{table*}

\subsection{Identified Degradation Factors}
\noindent\textbf{1. Unfavourable Light Condition}. There is a wide range of factors that could cause brightness variations in automotive camera images, such as the different times of the day, the inconsistency of the exposure, the dynamic range of the camera sensors and imperfection of the lens \cite{tan2021night,zhou2022lednet,ceccarelli2022rgb}. Moreover, there exist complex light conditions for real-world AAD, especially at nighttime with multiple light sources such as headlights, streetlights, etc. In the literature, there is no unified definition of unfavourable light conditions; therefore, this work proposes an intuitive definition of four types of unfavourable light conditions: \textbf{extreme light, night light, low light and strong light (from darker to brighter)}, as shown in Tab.~\ref{light}.
Each one of these lighting conditions is treated separately and with three levels of severity in our study.  

\begin{table}[]
\caption{Proposed definition of unfavourable light, including low-light, night light and extreme light, classified by illumination level and glare intensity.}
\label{light}
\begin{tabular}{l|l|l}
\hline
\rowcolor[HTML]{9B9B9B} 
\multicolumn{1}{c|}{\cellcolor[HTML]{9B9B9B}{\color[HTML]{FFFFFF} \textbf{Low Light}}} &
  \multicolumn{1}{c|}{\cellcolor[HTML]{9B9B9B}{\color[HTML]{FFFFFF} \textbf{Night Light}}} &
  \multicolumn{1}{c}{\cellcolor[HTML]{9B9B9B}{\color[HTML]{FFFFFF} \textbf{Extreme Light}}} \\ \hline
\rowcolor[HTML]{EFEFEF} 
\begin{tabular}[c]{@{}l@{}}Uniform darkening \\ of images (without \\ glare and flare)\end{tabular} &
  \begin{tabular}[c]{@{}l@{}}Nighttime urban driving \\ area with vibrant street \\lights (with glare and flare)\end{tabular} &
  \begin{tabular}[c]{@{}l@{}}Darker illumination \\  compared to low \\ light and night light\end{tabular} \\ \hline
\end{tabular}
\end{table}

\noindent\textbf{Novel unfavourable light models}. 
In the context of diverse light levels encountered in real-world AAD scenarios, especially during the nighttime with the presence of multiple light sources, we introduce a unifying definition of unfavourable light (see Tab. \ref{light}) and a novel approach tailored to each specific lighting condition. This distinct categorization ensures a nuanced and accurate representation of various real-world lighting conditions (See Fig. 1). The complex real-world conditions not only result in unfavourable illumination, limiting the camera's ability to capture detailed information but also critically impact perception accuracy, as underscored in prior research \cite{lin2020gan}.  
This paper abandons the commonly used gamma correction \cite{lv2018mbllen}, which is effective in generating different brightness for indoor static objects but inadequate for simulating real-world dark conditions characterized by uneven light distribution and pixel saturation. Therefore, the low-light images are generated via $I^{1}_{low} = EZ_{dce}(I, \theta )$ to keep the saturated pixels while darkening the other areas. $EZ_{dce}$ is the reversed curve estimation-based image darkening method naming EC-Zero-DCE \cite{zhou2022lednet}, $\theta $ is the parameters of the model. For nighttime images, the darkening process alone cannot produce the flare and glare features that are invariably present in nighttime photographs. Thus, we generate the night light images $N$ using cycleGAN ($I^{1}_{night} = GAN_{cyc}(I)$)  with the subsequent cycle consistency loss function $ L_{cyc}(G,F) = E_{I\sim p_{data}(I)}\| F(G(I)) - I \|_{1} + E_{N\sim p_{data}(N)}\| G(F(N)) - N \|_{1}$ \cite{zhu2017unpaired}. $G()$ and $F()$ are the generators for clean-to-night and night-to-clean, respectively. Furthermore, the night light model is insufficient for extremely dark images since there remains a tiny percentage of images with lighting that is essentially almost unchanged from daylight. Therefore, extreme light images are generated by the compound of the above two models with the equation: $$I^{1}_{extreme} = EZ_{dce}(GAN_{cyc}(I), \theta )$$.

\noindent\textbf{2. Adverse Weather}. The effect of adverse weather on automated driving and cameras has been studied broadly \cite{8666747,michaelis2019benchmarking,brophy2023review}. These natural phenomena can reduce the quality of the captured images, hence causing potential safety risks for AAD systems. The impact on the image quality can be worse depending on the intensity of the phenomena, and it opens questions regarding the ability of vision-based automated vehicles to cope with not ideal environmental situations. This work considers adverse weather (fog, rain) with a newly introduced snow model described below.

\noindent\textbf{Novel Proposed Snow Model.}
\begin{figure}[t!]
\centering
\includegraphics[width=.5\textwidth]{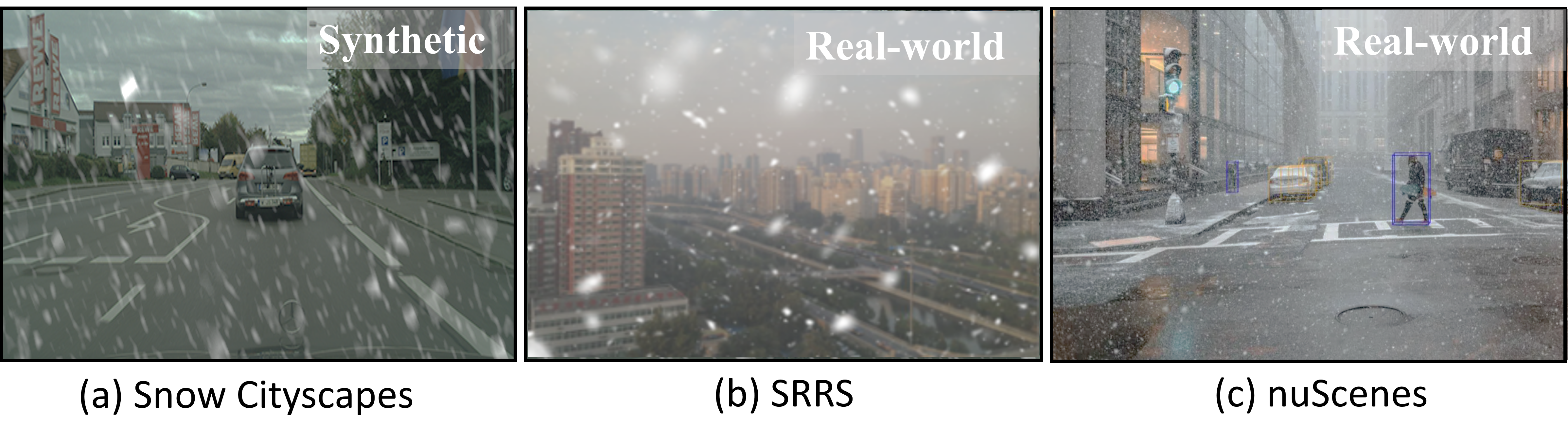}
\caption{The visual examples of the different snow images, from left to right: (a) Synthetic extreme snow without veiling effect from snow cityscapes \cite{zhang2021deep} (b) and (c) are real-world snow images from \cite{chen2020jstasr,caesar2020nuscenes}, where the clear veiling effect with some mist-like mask can be observed.}
\label{snow-veiling}
\vspace{-5pt}
\end{figure}
Due to the varied shape of the snowflakes, most of the existing snow simulation methods simply use PhotoShop to augment the snow layer into the clean image layer, such as the Snow Cityscapes dataset \cite{zhang2021deep} and the Snow 100K dataset \cite{liu2018desnownet}. 
However, the existing models neglect the veiling effect (haze or mist-like effect) when considering Koschmieder's theory of image degradation caused by light scattering and absorption \cite{koschmieder1924theorie} (see Fig. \ref{snow-veiling}). Although the veiling effect has been considered in some synthetic snow datasets (e.g. Jstars \cite{chen2020jstasr}, CSD \cite{chen2021all}), these are not specific to the driving scenes. Therefore, in this research, we improve the snow model with the added veiling effect to synthesise the snow dataset $I^{2'}_{snow}$ with the following equation:

\begin{equation}
\begin{aligned}
I^{2}_{snow} = z(x)\  \odot A(x) + I(x)\  \odot (1-z(x)) \\
I^{2'}_{snow} = I^{2}_{snow} * e^-\beta d(x) + A(x) (1- e^-\beta d(x)) ,
\end{aligned}
\end{equation}

\noindent where $I^{2}_{snow}$ are the non-veiling effect snowy images. $\  \odot$ is the element-wise multiplication, $A(x)$ represents the chromatic aberration map, and $z(x)$ is the independent snow mask. $e^-\beta d(x)$ is the median transmission map. $\beta$, and $d(x)$ are the scattering coefficient and the distance of the object to the camera, respectively.

\begin{figure}[t!]
\centering
\includegraphics[width=.5\textwidth]{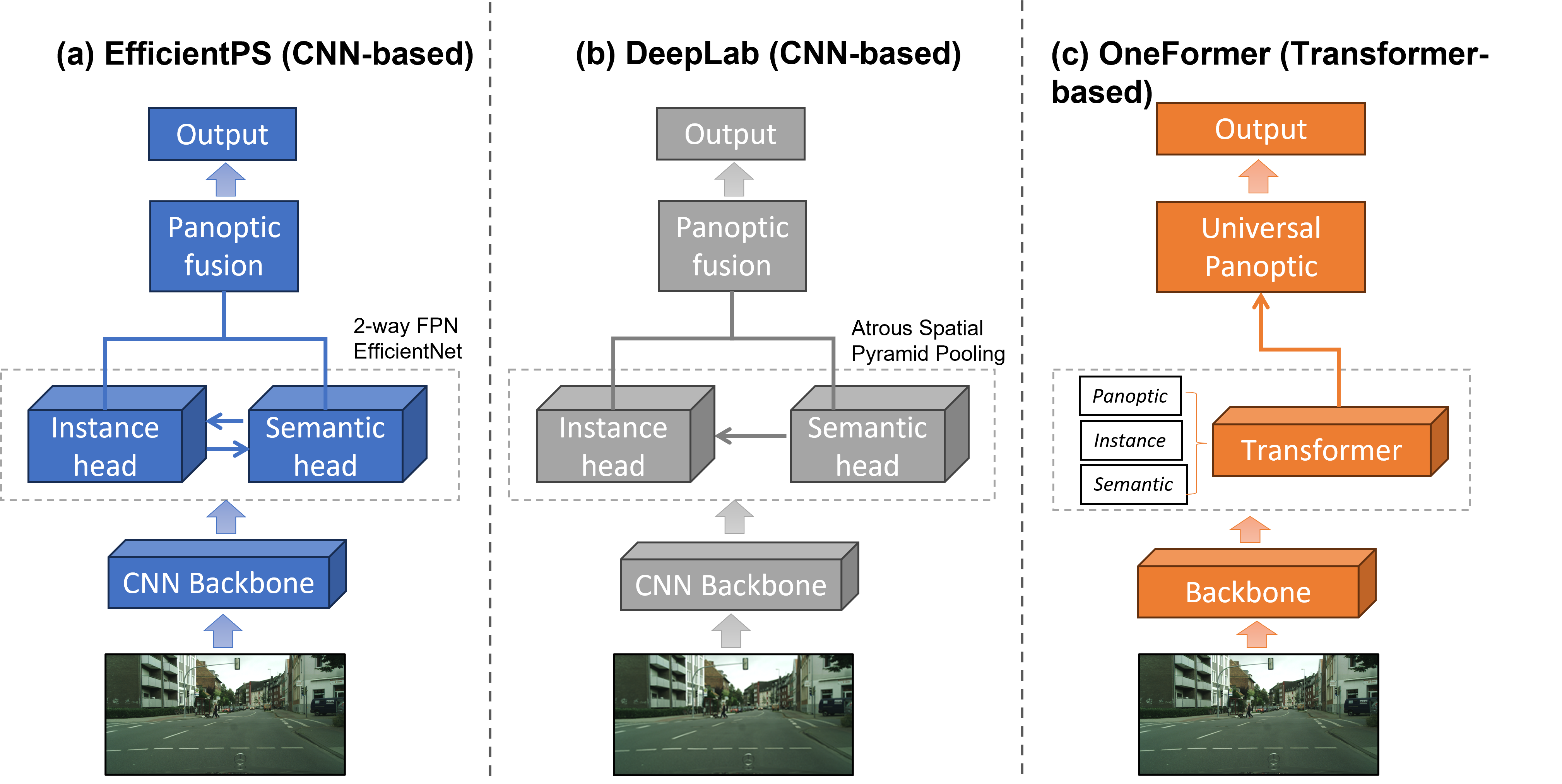}
\caption{The overview of the selected panoptic segmentation models, (a) EffcientPS \cite{mohan2021efficientps} (b) DeepLab \cite{cheng2020panoptic} (c) Oneformer\cite{jain2023oneformer}.}
\label{panoptic_models}
\vspace{-5pt}
\end{figure}


\noindent\textbf{3. Optic Obstruction}. Optic obstructions are likely to occur under certain circumstances with lens occlusions, such as droplets on the lens and mud obstructions. These occlusions will result in information loss or distortion of some portion of the frames, causing, for example, a performance drop in object detection~\cite{li2022analysis}.

\noindent\textbf{4. ISP Failure}. Raw images contain rich and unprocessed data information, and some deep learning techniques have been created and optimised to handle ISP-proceeded RGB images~\cite{chan2023raw}. 
Errors in the ISP imaging process, for example, demosaicing, compression or sharpening, can lead to inaccurate information. Moreover, using different Bayer filters can introduce variations in colour representation, subsequently impacting the outcomes of deep learning algorithms. Therefore, this research examines the impact of compression, over-sharpening, varied Bayer filters and demosaicing in detail.

\noindent\textbf{5. Internal Sensor Noises}. The exploration of diverse sensor noise factors is a fundamental endeavour in the domain of image corruption analysis, a focus extensively examined in robustness research studies \cite{dong2023benchmarking,kamann2021benchmarking}. These noise factors stem from multifaceted origins, ranging from fluctuations in external light conditions to internal camera lens intricacies, variations in ISP processes, and temporal changes over time. 

\noindent\textbf{6. Vehicle Motion.} 
In AAD, vehicle motion and mounting vibrations can result in motion blur and unfocused blur impacting the quality of captured images, which introduces a significant challenge \cite{li2019event}. Mounting vibrations, arising from various sources such as road irregularities, vehicle movements, or external factors, have a profound effect on the focus stability of onboard cameras \cite{wang2023joint}. This translates to the loss of critical high-frequency information, such as the detailed boundary delineation of vehicles and pedestrians. This information is indispensable for AAD systems, enabling accurate object recognition and tracking. 



\section{Proposed Pipeline}
The proposed unifying degradation impact pipeline, shown in Fig.~\ref{pipline}, consists of three main steps: synthesis of noisy images (see previous Section) using the selected dataset, panoptic segmentation, and impact evaluation. This chapter gives details about the dataset, selected panoptic models and evaluation metrics.
\subsection{Dataset}
The Cityscapes dataset \cite{cordts2016cityscapes} is used as the clean dataset for our synthetic data. The Cityscapes dataset comprises high-resolution daytime images sourced from 50 European cities, meticulously annotated with 19 classes at the pixel level and 30 classes at the instance level. The dataset is chosen to generate the D-Cityscapes+ for the following reasons: 1) it is one of the most commonly used driving datasets; 2) it is captured with high quality during the day with less noise that may influence the synthetic process; 3) the dataset contains labels for both instance segmentation and semantic segmentation, along with existing panoptic models that are trained on this dataset. To maintain consistency and expedite comparisons, all 500 images in the dataset are employed for validation. For streamlined analysis, the images were resized to a standard resolution of 1024$\times$512 using bicubic interpolation, ensuring both efficiency and uniformity throughout the evaluation process.
\subsection{Selected Panoptic Segmentation Models.} Based on the size of the network and the panoptic quality (PQ) on the clean Cityscapes dataset, three state-of-the-art panoptic segmentation models are selected, which are the Panoptic Deeplab \cite{cheng2020panoptic}, EfficientPS \cite{mohan2021efficientps} and Oneformer \cite{jain2023oneformer}, Fig.~\ref{panoptic_models}. Amongst the different chosen models, benefiting from the Efficient Net \cite{tan2019efficientnet} backbone, EfficeintNet has the lowest number of parameters (40.9M) compared with the highest (372M) with relatively good performance, i.e. PQ=62.8. In terms of the panoptic deeplab \cite{cheng2020panoptic}, it uses Atrous Spatial Pyramid Pooling (ASPP) which is one of the most commonly used architectures in segmentation tasks \cite{kamann2021benchmarking}. This can benefit the model with multiscale feature learning ability, a larger perceptive field, relatively fewer parameters (46.7M) and good performance (QP=63). The Oneformer \cite{jain2023oneformer} leverages the long-range relationship learning ability via the Swin transformer \cite{liu2021swin} and CovNet \cite{liu2022convnet} architecture, which archives the best performance in terms of the PQ (70.1) with the largest number of parameters (220-372M). Six different backbones were used during the experiment, three for the Panoptic Deeplab (i.e. ResNet, Xception Net, and HR48 Net) and three for the Oneformer (i.e. Swin-L, ConvNeXt-L, ConvNeXt-XL) to compare their robustness.

\subsection{Evaluation Metrics}
This research evaluates the degradation impact from three perspectives: synthetically generated noisy image quality evaluation, image evaluation using panoptic segmentation-based perception, and the correlation between them.

\subsubsection{Image Quality Analysis}
To analyse the image quality reduction due to the selected noise factors, 8 image quality metrics are selected with a wide coverage of the image features from both the spatial domain (i.e. local mean, local contrast, edge gradients, chrominance) and the frequency domain (i.e. Fourier- and Wavelet-based) \cite{gummadi2023correlating}. They encompassed both full-reference metrics (i.e. PSNR \cite{PSNR}, SSIM \cite{zhou2004SSIM}, FID \cite{heusel2018FID}, LPIPS \cite{zhang2018unreasonable}, CW-SSIM \cite{sampat2009CW-SSIM}, FSIM\cite{zhang2011FSIM}) and no-reference metrics (i.e.
BRISQUE \cite{mittal2012BRISQUE}, NIQE \cite{mittal2013NIQE}). 
The image signal-to-noise ratio and the structural information degradation are quantified by PSNR and SSIM, respectively. Lower PSNR and SSIM scores indicate a higher impact on the noise factors, causing loss of information. The CW-SSIM and FSIM evaluate the degradation from a frequency perspective; higher values indicate less degradation in the frequency domain. 
FID calculates the feature distance between the ground truth image and the generated ones. In addition, the NIQE and LPIPS scores evaluate perceptual differences between the images; the smaller these values represent, the better the naturalness of the synthetic images. 
BRISQUE uses a trained Support vector machine (SVM) to compute a quality score; a lower score indicates better image quality. 

\begin{figure}[t!]
\centering
\includegraphics[width=.5\textwidth]{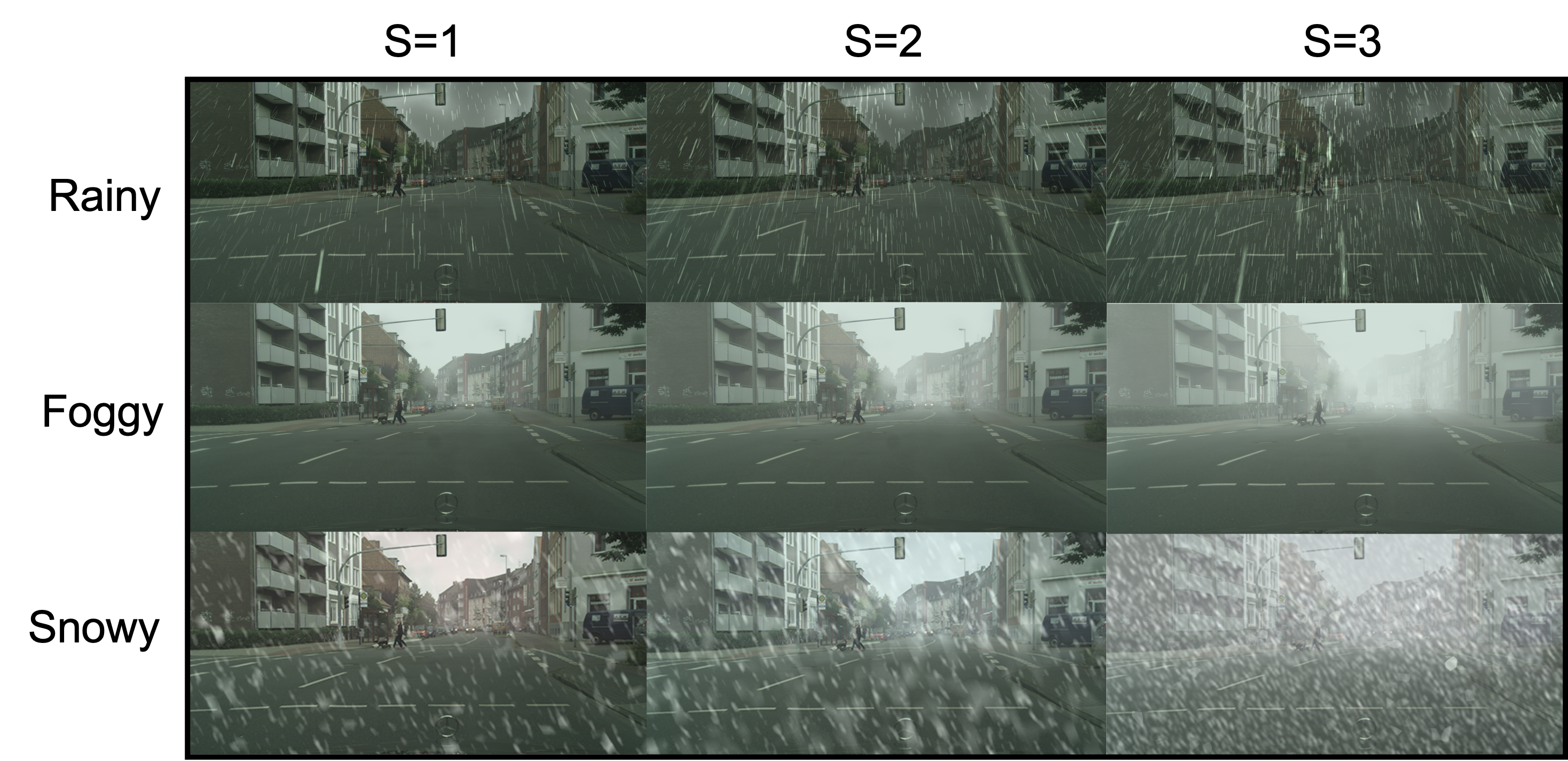}
\caption{The visual examples of the different severity levels of the adverse weather conditions, from 1st to 3rd rows: rainy, foggy and snowy conditions, and from left to right: light, medium and heavy.}
\label{multi-severity-weather}
\vspace{-5pt}
\end{figure}


\subsubsection{Panoptic Robustness Evaluation Metrics} 
To assess how panoptic models perform against the various noise models, the widely used panoptic quality (PQ) measure is adopted. The index PQ serves as the ideal indicator for this research since it contains both object-level information and fine-grained pixel-level information. Specifically, PQ is the product of segmentation quality ($SQ$) and recognition quality ($RQ$).
$SQ$ only considers results to be a match when the overlap between the ground truth and the prediction $IOU$ value is above 0.5. 
$RQ$ calculates the true positive (TP), false positive (FP), and false negative (FN) to get the precision and recall. 
The larger the PQ values, the better the quality. The PQ can be formulated as in Eq.~\ref{PQ}.

\begin{equation}
\small
PQ\  =\  \frac{\sum^{g}_{s} IOU(s,g)}{\left| TP\right|  } \times \  \frac{\left| TP\right|  }{\left| TP\right|  +\frac{1}{2} \left| FP\right|  +\frac{1}{2} \left| FN\right|  } \  \  
\label{PQ}
\end{equation}
\noindent where $s$ represents the segmentation results, $g$ represents the ground truth, and $(s,g)\in TP$. For a set of $N$ images, the average PQ is calculated as: $aPQ = {\textstyle \sum_{i=1}^{N}} PQ_{i}/N $. 


\subsubsection{Correlation Metrics} The correlation between the 8 image quality index and the panoptic quality is calculated using Pearson's linear correlation coefficient (PLCC) and Spearman’s rank correlation coefficient (SRCC) \cite{gummadi2023correlating}. The main difference between PLCC and SRCC is that the former is mainly based on the value, while the latter is based on the rank of each value.


\section{Experimental Results and Analysis}
This section gives the qualitative and quantitative experimental results and analysis in terms of the image quality of the generated noisy images, the impact of the noise factors in panoptic segmentation and the robustness between different panoptic segmentation models under different degradation levels.


\subsection{Generated Noisy Images (D-Cityscapes+)}
\textbf{Visual quality}
The visual results of the proposed D-Cityscapes+ are shown in Fig.~\ref{D-Cityscapes+}, where 19 types of noise factors are considered. Fig.~\ref{realism} compares the synthesized multi-weather (i.e. fog, snow, and rain) to state-of-the-art robustness benchmarking research \cite{dong2023benchmarking} using the Cityscape dataset. Fig.~\ref{multi-severity-weather} displays the visual results of the multi-weather at different severity levels (s=1, 2, and 3, where 3 corresponds to the most severe conditions). From these visual results, it can be inferred that 
compared to real-world captured datasets, the newly created dataset shows a better coverage of extreme weather conditions, as they are rarely seen and difficult to capture in the real world. 
In addition, (b) and (c) from Fig. \ref{snow-veiling} indicate the veiling effect (i.e. mist-like background and added blur in further distance) can be observed in the real world, while Snow Cityscapes (see (a) from Fig.~\ref{snow-veiling})
do not obey the same observation  (i.e. clear background in the distance) \cite {zhang2021deep}, which does, therefore, not obey the characteristics of the images captured by the automotive camera in AAD.
Therefore, in this work, the snow veiling effect is modelled and added with different severity levels according to the different severity levels of the precipitation, Fig.~\ref{multi-severity-weather}. Especially under medium- to heavy snow conditions, the objects in the further distance show a mist-like and blurred effect due to the clustering of snowflakes. (b) (c) compared with (a), it is not simulated in (a); therefore, this paper adds this effect to the snow model.

\begin{figure*}[t!]
\centering
\includegraphics[width= 0.8\textwidth]{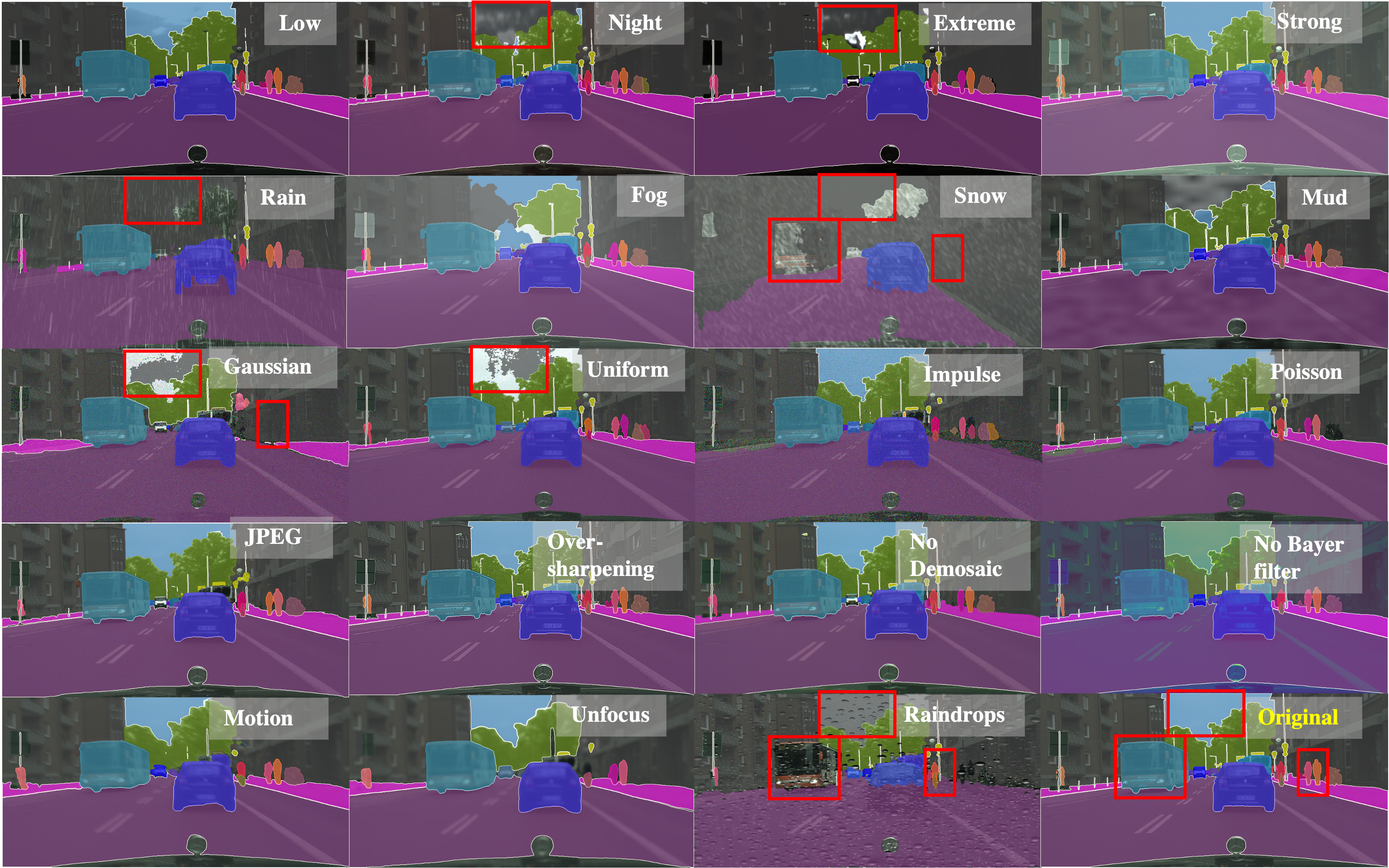}
\caption{The overview of the panoptic segmentation using the best performing Oneformer \cite{jain2023oneformer} \textit{Errors are shown in the red bounding box}.}
\label{Visual_Oneformer}
\vspace{-5pt}
\end{figure*}

\begin{table*}
    \centering
    \caption{The panoptic segmentation quantitative results using Oneformer on D-Cityscapes+ with severity $s=3$, note that $PQ_s$ means the average PQ value at severity $s$, $PQ_s^v$ means the variance PQ at severity $s$.}
    \begin{tabular}{cccccccccc}
    
        \toprule[1pt]
        \multicolumn{1}{c|}{\textbf{Metrics}} & 
        \multicolumn{1}{c|}{\textbf{Clean}} & 
        \multicolumn{2}{c|}{\textbf{\makecell{light level}}} & 
        \multicolumn{3}{c|}{\textbf{\makecell{Weather}}} & 
        \multicolumn{2}{c|}{\textbf{\makecell{Lens}}} & 
        \multicolumn{1}{c}{\textbf{\makecell{ISP}}} \\
        \hline
    
        \multicolumn{1}{c|}{} & 
        \multicolumn{1}{c|}{} & 
        \multicolumn{1}{c}{Dark} & 
        \multicolumn{1}{c|}{Strong} & 
        \multicolumn{1}{c}{Rainy} & 
        \multicolumn{1}{c}{Foggy} & 
        \multicolumn{1}{c|}{Snowy} & 
        \multicolumn{1}{c}{mud obstructions} & 
        \multicolumn{1}{c|}{Rain Drops} & 
        \multicolumn{1}{c}{Compression}
        \\
        \hline
        \multicolumn{1}{c|}{\textbf{\textit{Swin-L† ($PQ_1$)}}} & 
        \multicolumn{1}{c|}{\textbf{67.2}} & 
        \multicolumn{1}{c|}{63.4} & 
        \multicolumn{1}{c|}{\textcolor{blue}{66.1}} & 
        \multicolumn{1}{c|}{58.5} & 
        \multicolumn{1}{c|}{64.0} & 
        \multicolumn{1}{c|}{48.8} & 
        \multicolumn{1}{c|}{65.4} & 
        \multicolumn{1}{c|}{\textcolor{red}{34.9}} & 
        \multicolumn{1}{c}{58.2}         
        \\

        \multicolumn{1}{c|}{\textbf{\textit{ConvNeXt-L† ($PQ_1$)}}} & 
        \multicolumn{1}{c|}{\textbf{68.5}} & 
        \multicolumn{1}{c|}{64.2} & 
        \multicolumn{1}{c|}{\textcolor{blue}{67.0}} & 
        \multicolumn{1}{c|}{60.9} & 
        \multicolumn{1}{c|}{65.0} & 
        \multicolumn{1}{c|}{50.1} & 
        \multicolumn{1}{c|}{66.7} & 
        \multicolumn{1}{c|}{\textcolor{red}{38.5}} & 
        \multicolumn{1}{c}{59.8} 
        \\      

        \hline        
        \multicolumn{1}{c|}{\textbf{\textit{Swin-L† ($PQ_2$)}}} & 
        \multicolumn{1}{c|}{\textbf{67.2}} & 
        \multicolumn{1}{c|}{60.2} & 
        \multicolumn{1}{c|}{\textcolor{blue}{64.5}} & 
        \multicolumn{1}{c|}{50.2} & 
        \multicolumn{1}{c|}{61.6} & 
        \multicolumn{1}{c|}{37.5} & 
        \multicolumn{1}{c|}{63.0} & 
        \multicolumn{1}{c|}{47.7} & 
        \multicolumn{1}{c}{54.3}         
        \\

        \multicolumn{1}{c|}{\textbf{\textit{ConvNeXt-L† ($PQ_2$)}}} & 
        \multicolumn{1}{c|}{\textbf{68.5}} & 
        \multicolumn{1}{c|}{61.5} & 
            \multicolumn{1}{c|}{\textcolor{blue}{66.4}} & 
        \multicolumn{1}{c|}{54.4} & 
        \multicolumn{1}{c|}{63.1} & 
        \multicolumn{1}{c|}{38.4} & 
        \multicolumn{1}{c|}{65.5} & 
        \multicolumn{1}{c|}{42.0} & 
        \multicolumn{1}{c}{53.6} 
        \\      

        \hline        

        \multicolumn{1}{c|}{\textbf{\textit{Swin-L† ($PQ_3$)}}} & 
        \multicolumn{1}{c|}{\textbf{67.2}} & 
        \multicolumn{1}{c|}{52.5} & 
        \multicolumn{1}{c|}{\textcolor{blue}{62.6}} & 
        \multicolumn{1}{c|}{36.1} & 
        \multicolumn{1}{c|}{56.9} & 
        \multicolumn{1}{c|}{21.5} & 
        \multicolumn{1}{c|}{60.0} & 
        \multicolumn{1}{c|}{36.6} & 
        \multicolumn{1}{c}{46.6}         
        \\

        \multicolumn{1}{c|}{\textbf{\textit{ConvNeXt-L† ($PQ_3$)}}} & 
        \multicolumn{1}{c|}{\textbf{68.5}} & 
        \multicolumn{1}{c|}{53.5} & 
        \multicolumn{1}{c|}{63.9} & 
        \multicolumn{1}{c|}{36.6} & 
        \multicolumn{1}{c|}{59.3} & 
        \multicolumn{1}{c|}{20.5} & 
        \multicolumn{1}{c|}{\textcolor{blue}{64.7}} & 
        \multicolumn{1}{c|}{31.6} & 
        \multicolumn{1}{c}{43.8} 
        \\      


        \midrule[1pt]
        \multicolumn{1}{c|}{\textbf{Metrics}} &
        \multicolumn{4}{c|}{\textbf{Sensor Noises}} & 
        \multicolumn{2}{c|}{\textbf{Blur}} & 
        \multicolumn{3}{c}{\textbf{ISP}}
        \\
        \hline

        
        \multicolumn{1}{c|}{} & 
        \multicolumn{1}{c}{Gaussian} & 
        \multicolumn{1}{c}{Uniform} & 
        \multicolumn{1}{c}{Impulse} & 
        \multicolumn{1}{c|}{Poisson} & 
        \multicolumn{1}{c}{Unfocus} & 
        \multicolumn{1}{c|}{Motion} & 
        \multicolumn{1}{c}{O-Sharp} & 
        \multicolumn{1}{c}{N-Dem} & 
        \multicolumn{1}{c}{\textbf{Colour-D}}
        \\
        \hline
        
        \multicolumn{1}{c|}{\textbf{\textit{Swin-L†($PQ_1$)}}} & 
        \multicolumn{1}{c|}{44.8} & 
        \multicolumn{1}{c|}{54.7} & 
        \multicolumn{1}{c|}{59.4} & 
        \multicolumn{1}{c|}{59.7} & 
        \multicolumn{1}{c|}{51.7} & 
        \multicolumn{1}{c|}{54.4} & 
        \multicolumn{1}{c|}{65.5} & 
        \multicolumn{1}{c|}{61.4} & 
        \multicolumn{1}{c}{61.3} 
        \\

        \multicolumn{1}{c|}{\textbf{\textit{ConvNeXt-L† ($PQ_1$)}}} & 
        \multicolumn{1}{c|}{48.4} & 
        \multicolumn{1}{c|}{57.5} & 
        \multicolumn{1}{c|}{58.0} & 
        \multicolumn{1}{c|}{62.5} & 
        \multicolumn{1}{c|}{52.6} & 
        \multicolumn{1}{c|}{55.7} & 
        \multicolumn{1}{c|}{65.5} & 
        \multicolumn{1}{c|}{61.0} & 
        \multicolumn{1}{c}{62.9}
        \\        

        \hline
        
        \multicolumn{1}{c|}{\textbf{\textit{Swin-L†($PQ_2$)}}} & 
        \multicolumn{1}{c|}{\textcolor{red}{28.2}} & 
        \multicolumn{1}{c|}{50.2} & 
        \multicolumn{1}{c|}{47.3} & 
        \multicolumn{1}{c|}{54.1} & 
        \multicolumn{1}{c|}{37.5} & 
        \multicolumn{1}{c|}{38.6} & 
        \multicolumn{1}{c|}{\textcolor{blue}{63.1}} & 
        \multicolumn{1}{c|}{61.4} & 
        \multicolumn{1}{c}{61.3} 
        \\

        \multicolumn{1}{c|}{\textbf{\textit{ConvNeXt-L† ($PQ_2$)}}} & 
        \multicolumn{1}{c|}{\textcolor{red}{32.4}} & 
        \multicolumn{1}{c|}{54.0} & 
        \multicolumn{1}{c|}{44.6} & 
        \multicolumn{1}{c|}{57.0} & 
        \multicolumn{1}{c|}{38.6} & 
        \multicolumn{1}{c|}{39.4} & 
        \multicolumn{1}{c|}{63.6} & 
        \multicolumn{1}{c|}{61.0} & 
        \multicolumn{1}{c}{62.9}
        \\        

        \hline
        \multicolumn{1}{c|}{\textbf{\textit{Swin-L†($PQ_3$)}}} & 
        \multicolumn{1}{c|}{\textcolor{red}{7.4}} & 
        \multicolumn{1}{c|}{47.4} & 
        \multicolumn{1}{c|}{16.7} & 
        \multicolumn{1}{c|}{49.0} & 
        \multicolumn{1}{c|}{25.2} & 
        \multicolumn{1}{c|}{27.8} & 
        \multicolumn{1}{c|}{61.7} & 
        \multicolumn{1}{c|}{61.4} & 
        \multicolumn{1}{c}{61.3} 
        \\

        \multicolumn{1}{c|}{\textbf{\textit{ConvNeXt-L† ($PQ_3$)}}} & 
        \multicolumn{1}{c|}{\textcolor{red}{11.5}} & 
        \multicolumn{1}{c|}{50.7} & 
        \multicolumn{1}{c|}{22.7} & 
        \multicolumn{1}{c|}{52.2} & 
        \multicolumn{1}{c|}{25.6} & 
        \multicolumn{1}{c|}{28.6} & 
        \multicolumn{1}{c|}{62.2} & 
        \multicolumn{1}{c|}{61.0} & 
        \multicolumn{1}{c}{62.9}
        \\        

        
        \bottomrule[1pt]
    \end{tabular}
    \label{PQ_Onformer}
\end{table*}

\textbf{Impact of the noise models and Image Quality}
\label{par:DF_Impact_On_IQ} 
The quantitative results of eight image quality metrics applied to data under all the noise factors are shown in Tab. 1 from the suppl. material. As can be observed in the Table, the synthetic noise models create frames with quality degrading according to the noise severity level. Unfavourable light conditions, however, deviate from the pattern except for the FID score, with a worse image quality for low light than night light. The reason for this discrepancy can be attributed to the fact that night light images tend to retain more local light from different light sources. On the other hand, low-light images have an even distribution of low illumination conditions. 
Another exception also exists for the mud obstructions, with a slightly smaller PSNR value when s=1 compared with when s=3. This result might be due to the generation process of the mud obstructions, where a bigger kernel size indicates more severe conditions, while the number of mud obstructions dynamically reduces to adapt to the image size. Therefore, this generation process results in the PSNR being more sensitive to the number than the size of the mud obstructions. 

As for the quality indicated by NIQE and BRISQUE, it is found that they do not show the same trend when increasing the severity levels, however, they are sensitive to noise and artefacts. This may be because NIQE compares the feature distribution in the given image with a pre-computed natural image distribution, which sometimes differs much between the randomly generated noise. BRISQUE instead uses regression-based scoring for feature learning, and this process sometimes results in a different trend from NIQE. As for the frequency-based image quality, CW-SSIM and FSIM generally align with the trend of SSIM. With the potential reason that the frequency and spatial values can indicate the image's structural information. Overall, FID consistently aligns with the trend of increasing degradation severity across all factors, making it the most versatile metric to use. However, specific metrics, like BRISQUE and PSNR, can offer valuable insights and context for understanding the impact of particular noise factors on image quality, as they are sensitive to slight changes.

\subsection{Panoptic Segmentation Results}
\textbf{Impact of noise models on Panoptic Segmentation}
The quantitative and visual results using the Oneformer \cite{jain2023oneformer} under 19 types of noise models are illustrated in Tab.~\ref{PQ_Onformer} and Fig.~\ref{Visual_Oneformer}. \textit{see suppl. material for more details).}
Fig.~\ref{Visual_Oneformer} shows the same frame with noise models (at the highest severity level) and the panoptic segmentation results. With the increasing severity levels, all degradation factors (except the raindrop occlusion at s=2,3) show decreasing PQ values. 
Amongst the noise factors, the intensity, distortion, and scales of the artefacts generated in the frames will directly influence the panoptic segmentation performance. For example, the Gaussian noise and raindrop occlusion are the most influencing factors, while strong light, mud obstructions, and over-sharpening show the smallest impact. Overall, under all degradation factors, the ConvNeXt-L and ConvNeXt-XL perform better compared with the Swin-L backbone, indicating increased noise robustness in these architecture designs. Specifically, the ConvNeXt-L and ConvNeXt-XL show similar PQ, with ConvNeXt-L taking less processing time. 
ConvNeXt-XL is better at processing the compression and greyscale data, while the ConvNeXt-L largely surpasses the ConvNeXt-XL under Gaussian noise and impulse noise conditions.

To make the comparison more meaningful, we analyse the different panoptic performances within the same ID categories. \textit{1) Light level.} As can be seen from the dark light results, the uniform darkening of the illumination shows the least degradation, and the increasing dark level, and the dynamic exposure (e.g. glare and flare) can influence the panoptic segmentation performance. The categories of strong light show that too much illumination could also result in slightly reduced performance. \textit{2) Adverse weather.} With decreasing visibility, increasing intensity and bigger particles occurring during adverse weather conditions, the performance decreases (i.e. Snow\textless Rain \textless Fog) under all severity levels. \textit{3) Lens.} The mud obstructions in the used noise models show better panoptic segmentation performance compared with the image lens covered with droplets on the lens at the same severity levels, which might be caused by a distortion around the droplets on the lens. \textit{4) Sensor noise.} The Gaussian noise is one of the most impacting sensor noise factors in terms of the drop of PQ.
\textit{5) Blur.} The unfocus blur can result in slightly worse performance than motion blur at each severity level. \textit{6) ISP.} For the simulation of the ISP failures, compression shows the worst performance, while over-sharpening shows the least decrease in panoptic quality. textcolor{magenta}{Furthermore, the variance of the PQ ($vPQ$) can be seen in Tab. 3 from the suppl. material, the higher values in $vPQ$ for low-light or night light show worse stability for the deep-learning-based image simulation methods compared with the physical model-based ones.}
The investigation conducted in this study substantiates that within the framework of synthetic data, Gaussian noise significantly influences panoptic segmentation quality. Yet, the challenge lies in linking this specific noise distribution to real-world automotive scenarios. 

\subsection{Robustness of different panoptic segmentation models}
\begin{figure}[t!]
\centering
\includegraphics[width= 0.5\textwidth]{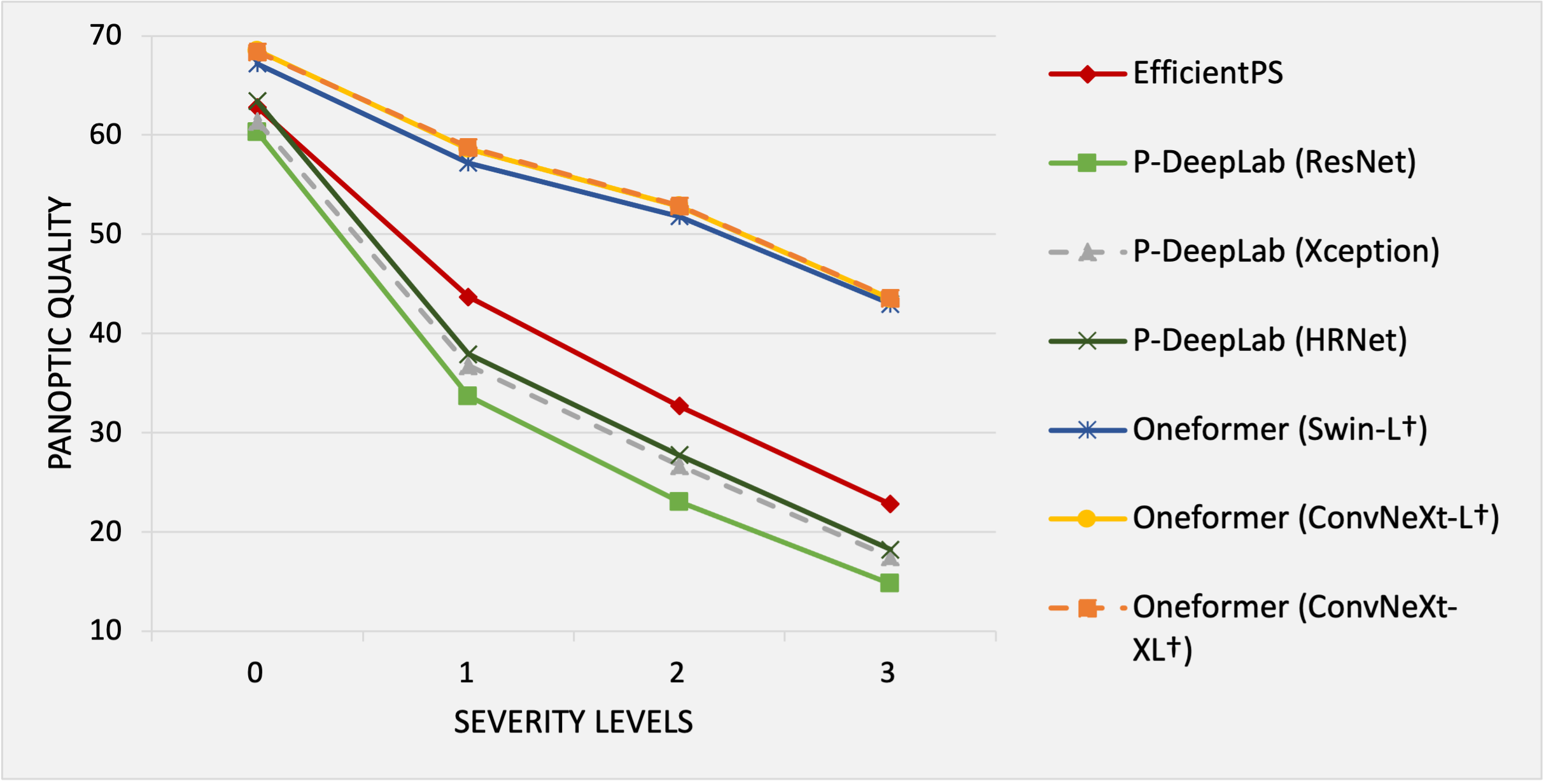}
\caption{The comparison of the panoptic segmentation results using 7 different backbones.}
\label{panoptic_backbones}
\vspace{-5pt}
\end{figure}

\begin{figure}[t!]
\centering
\includegraphics[width= 0.5\textwidth]{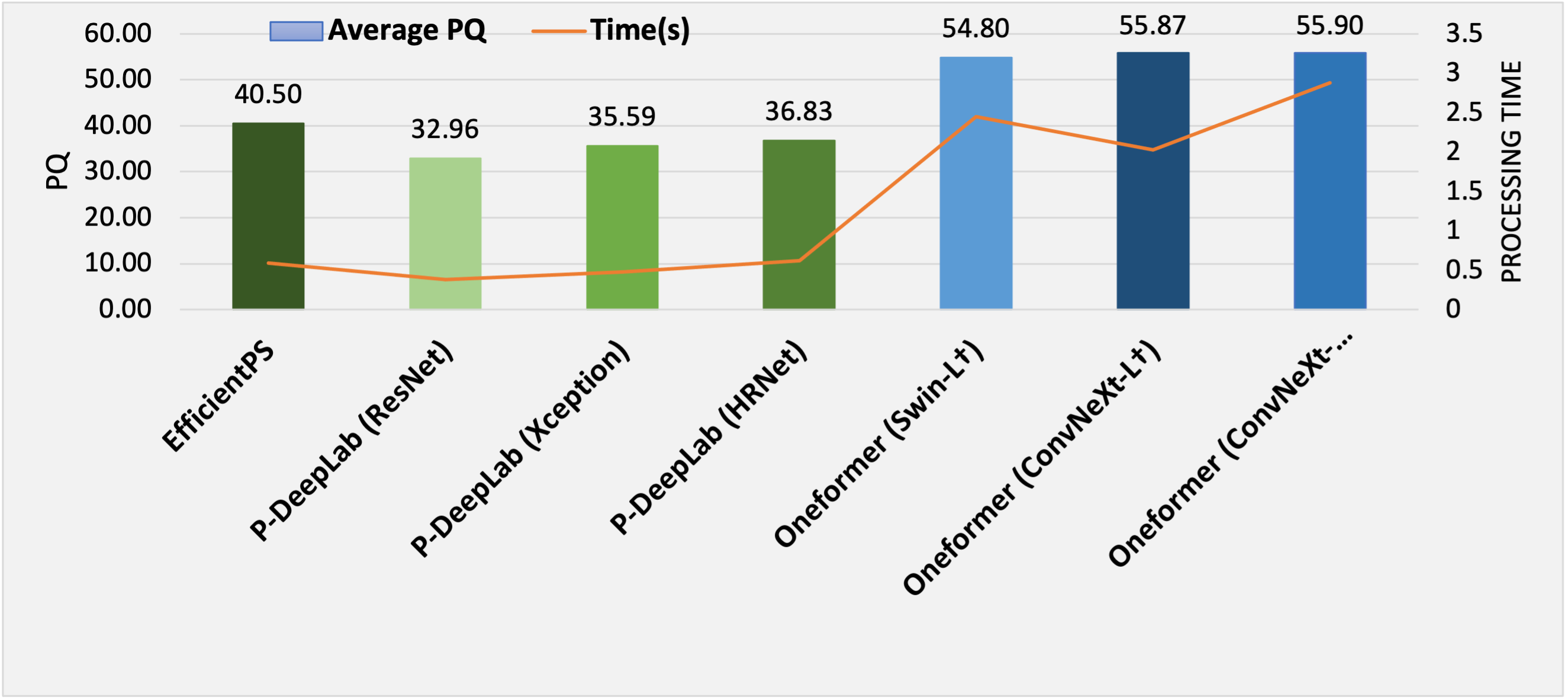}
\caption{The comparison results of the panoptic quality and time.}
\label{panoptic_time}
\vspace{-5pt}
\end{figure}

In the comparative study, three state-of-the-art panoptic segmentation models and various backbones have been evaluated, Fig.~\ref{panoptic_models}, to gauge their ability to handle noise factors~\cite{mohan2021efficientps,cheng2020panoptic,jain2023oneformer}. Figs.~\ref{panoptic_backbones}-\ref{panoptic_time} visually represent the comparison among these models in terms of the panoptic quality and speed (time per frame) under three different severity levels of the noise factors. The PQ values obtained across diverse degradation factors and backbones can be seen in Tab. 3-5 from the suppl. material. Notably, the presented analysis juxtaposes the robustness of CNN-based methods against ViT-based methods, revealing intriguing patterns in architectural efficiencies.

Surpassing its counterparts, Oneformer (pre-trained ConvNeXt-L and ConvNeXt-XL) emerges as the standout performer with an average PQ of 55.9, whereas Panoptic deeplab utilizing ResNet exhibits lower performance with an average PQ of 32.96. However, it is essential to note that higher PQ often corresponds to increased processing time and computational demand. 
Models with lower PQ values showcase quicker processing capabilities, suggesting a trend in current mainstream methods aiming for an equilibrium between performance and computational efficiency. Furthermore, the potential for enhanced robustness and generality of larger models becomes evident, especially with respect to internal sensor noise factors. For instance, the average PQ values in 4 different internal sensor noise factors under different severity levels for the Oneformer is much better (i.e. $aPQ_1$=55.7) compared with both P-Deeplab (i.e. $aPQ_1$=14.6)  and the EfficientPS (i.e. $aPQ_1$=28.1).

Delving into backbone architectures, the ConvNeXt-L, XL, a transformer-based CNN architecture, exhibits marginal superiority over conventional transformer backbones like Swin-L, far outpacing performance compared to traditional CNN backbones (ResNet, Xception, HRNet, and EfficientNet) despite having larger model parameters. Notably, the upgraded convolutional network, incorporating Swin Transformer architecture into the classic ResNet, even surpasses the pure Swin Transformer-based method, hinting at the efficacy of employing larger kernels and depthwise convolution invertible networks and training techniques such as pre-training and adaptive activation functions to bolster performance and robustness in current CNN models. Additionally, EfficientPS utilizing EfficientNet showcases commendable performance in both PQ and speed, outperforming the HRNet-based Panoptic deeplab, underscoring the potential for leveraging EfficientNet backbones in light-weight robust network architecture design. These findings illuminate critical directions for future network architecture design, emphasizing the need for a nuanced balance between performance, computational cost, and the strategic integration of innovative architectural components to fortify robustness in automated driving applications.


\begin{table*}[t] \centering
\caption{The overall correlation between the 8 image quality index to the PQ of EfficeintPS about the average PLCC and SRCC. }
\begin{tabular}{llllllllll}
\hline
\textbf{Model} & \textbf{Index} & \textbf{PSNR↑} & \textbf{SSIM↑} & \textbf{FID↓}                  & \textbf{LPIPS↓}                & \textbf{NIQE↓} & \textbf{CW-SSIM↑}             & \textbf{FSIM↑} & \textbf{BRISQUE↓} \\ \hline
              
 EfficientPS & \textbf{PLCC} & 0.807         & 0.948         & -0.923                        & {\color[HTML]{FE0000} -0.948} & -0.513        & {\color[HTML]{FE0000} 0.954} & 0.921         & -0.519           \\
 EfficientPS & \textbf{SRCC} & 0.800         & 0.933         & {\color[HTML]{FE0000} -0.967} & -0.867                        & -0.533        & {\color[HTML]{FE0000} 0.967} & 0.933         & -0.6             \\ \hline

Oneformer(ConNeXt-L) & \textbf{PLCC} & 0.813         & 0.961         & -0.947                       & {\color[HTML]{FE0000} -0.979} & -0.569        & {\color[HTML]{FE0000} 0.975} & 0.966         & -0.661           \\
Oneformer(ConNeXt-L) & \textbf{SRCC} & 0.800         & 0.933         & {\color[HTML]{FE0000} -0.967} & -0.933                        & -0.533        & {\color[HTML]{FE0000} 0.967} & 0.933  & -0.6             \\ \hline

\end{tabular}
\label{avg_correlation}
\end{table*}

\subsection{Correlation between the Image Quality index and Panoptic Quality index}
For the analysed 19 noise factors, the average correlation indexes (PLCC and SRCC) between panoptic segmentation performance (PQ) and the selected image quality metrics is reported in Tab.~\ref{avg_correlation}. The individual correlation indexes (for each type of noise) can be seen in Tab. 6-7 from the suppl. material. From the table, the panoptic models show a similar trend regarding the indexes. Specifically, the most correlated positive and negative indexes are CW-SSIM and LPIPS for PLCC, and CW-SSIM and FID for SRCC, respectively. These high correlation values indicate that the image quality index can be potentially used for predicting perception degradation. For example, for snowy conditions, PQ degrades from 51.2 to 18.3 when CW-SSIM degrade from 0.738 to 0.529. The degradation factors with the worst LPIPS and CW-SSIM scores (e.g. the Gaussian noise) also indicate the worst panoptic quality performance. In addition, the metrics capturing structural information of the image (e.g. SSIM, CW-SSIM) show a better average correlation with PQ. This correlation might be due to the structural information or features being important in the panoptic segmentation process, as the edge of objects for each instance should be predicted correctly to have higher PQ scores. 

However, the non-reference-based metrics (i.e. NIQE and BRISQUE) show low correlation values (i.e. around 0.5) with PQ. This low correlation may be due to the features learnt from the training images used for these metrics, being not specific to driving scenarios. As in previous studies, PSNR shows the worst correlation scores compared to all the reference-based metrics, meaning that PSNR cannot be used as a prediction factor for the performance decrease of deep learning-based AAD perception \cite{gummadi2023correlating}. For example, the noise factor with the lowest PSNR values is strong light when s=2 or 3; while the PQ values for strong light show the best panoptic segmentation quality. This correlation analysis has a substantial impact on the development of AAD because it successfully relates standard image quality measurements to reveal the panoptic segmentation performance.   

\section{Conclusion}
This study proposes a novel holistic evaluation framework to assess the robustness of perception in combination with the quality evaluation of sensor data, specifically camera data. The framework includes: (i) the injection of different types of noise factors with different severity levels (19 noise factors injected, each with 3 severity levels); (ii) the generation of an augmented and balanced noisy dataset, hereby named D-Cityscapes+, that might be used for further robustness studies; (iii) the assessment of variation in perception performance due to the noisy frames and different panoptic models (iv) the correlation between image quality and perception quality, to provide a set of guidelines and metrics that can be predictive of machine learning performance. 
Moreover, this work proposes two new improved noise models: (1) a snow model including reduced visibility; (2) extreme light models. 

This comprehensive evaluation aims to unify diverse degradation factors impacting automotive cameras within automated driving systems. The outcomes of the most influential factors guide certain noise factors (i.e. Gaussian noises and the droplets on the lens), among different corner cases, should be given priority since they pose the greatest hazard on perception (i.e. perception) in AAD. The results of better overall robustness in the ViT-based backbone architectures unveil critical insights for future architectural selections in the presence of noisy data. Through the proposed meticulous evaluation encompassing image and panoptic quality metrics, this work offers a nuanced understanding of noise factors, empowering stakeholders and designers of driving applications. 
The findings presented here aim to further explore the robustness of perception in general and in panoptic segmentation, specifically tailored for automated driving. Our benchmarking framework serves as a catalyst for advancing research endeavours, fostering the realization of a higher level of driving automation.



\bibliographystyle{IEEEtran}
\bibliography{egbib}

\clearpage

\end{document}